\documentclass[11pt]{article}

\usepackage{acl}

\usepackage{times}
\usepackage{latexsym}
\usepackage[T1]{fontenc}
\usepackage[utf8]{inputenc}
\usepackage{microtype}
\usepackage{graphicx}
\usepackage{booktabs}
\usepackage{multirow}
\usepackage{amsmath}
\usepackage{xcolor}
\usepackage{pifont}
\usepackage{enumitem}
\usepackage{url}
\usepackage{subcaption}

\usepackage{xspace}
\newcommand{\sys}{Ryze\xspace}

\title{\sys: Evidence-Enriched Data Synthesis from Biomedical Papers}

\author{Yeqi Huang\textsuperscript{1} \quad Yue Chen\textsuperscript{1} \quad Yanwei Ye\textsuperscript{1} \quad Guanhao Su\textsuperscript{1} \quad Luo Mai\textsuperscript{1} \\
  \textsuperscript{1}University of Edinburgh, United Kingdom \\ Demo: \url{https://ryze.12th.day} \\ Youtube Link: \url{https://youtu.be/5L2YShSaIQQ} \\ GitHub: \url{https://github.com/Chivier/Ryze}}

\begin{document}
\maketitle

\begin{abstract}
General-purpose VLMs remain unreliable for biomedical research because valid answers in scientific papers depend on evidence split across figures, tables, charts, captions, and referring text.
Existing post-training pipelines are bottlenecked by costly expert annotation and by synthetic data that drops this evidence structure.
We present \sys, a fully automated system that converts raw biomedical papers into an \emph{evidence-enriched} training set and a domain-specialized VLM\@.
\sys synthesizes QA pairs with complete supporting evidence (visual element, caption, extracted structure, and referring paragraphs), reduces layout and OCR errors via chart/table-aware extraction and LLM-based cleansing, and applies a progress-gated post-training strategy combining supervised fine-tuning with reinforcement learning.
Starting from Qwen3-VL-8B, \sys produces BioVLM-8B at under \$200, achieving 48.0\% weighted accuracy on LAB-Bench---outperforming the base model by +12.6 percentage points (pp) and surpassing GPT-5.2 by +3.8\,pp.
We release \sys as open source together with the trained BioVLM-8B model.
\end{abstract}

%% ============================================================
%%  SECTION 1: INTRODUCTION
%% ============================================================
\section{Introduction}

General-purpose vision-language models (VLMs) can read and reason over everyday images and text, but they remain unreliable for biology and biomedical research workflows that depend on precise interpretation of scientific papers, especially figures, charts, tables, captions, and the surrounding prose that explains them. Recent post-training methods (supervised fine-tuning and reinforcement learning) can inject domain capability into a base model, yet in practice the dominant bottleneck is still domain-specific training data: obtaining high-quality biomedical QA pairs typically requires costly expert annotation, and existing public datasets often fail to match the needs of targeted scientific tasks.

This paper addresses a concrete systems problem: \textbf{how to turn a corpus of raw biology/biomedical papers into a specialized VLM at low cost, without human annotation, while preserving strong evidence-grounded reasoning}. The central challenge is not simply to generate QA pairs, but to preserve the contextual evidence that makes scientific answers valid. In scientific papers, this evidence is often split across visual elements (e.g., axes, legends, and multi-row table headers) and the surrounding text that interprets them. When training examples drop these components, models tend to learn shallow cues or memorized patterns rather than evidence-grounded reasoning. 

While LLM-driven data synthesis has shown promise in general domains~\citep{wang-etal-2023-self-instruct,liu2023llava}, applying such methods naively to complex scientific PDFs risks producing hallucinated contexts.
Popular general-purpose tools—such as Meta Synthetic Data Kit~\citep{meta2025syntheticdatakit}, Marker~\citep{paruchuri2025marker}, and DeepSeek OCR v2~\citep{deepseek2025ocr}—often have poor data efficiency and struggle to produce sufficiently high-quality training data to improve models for biology and biomedical use cases. Further, prior biomedical-optimized VLMs such as LLaVA-Med~\citep{li2023llavamed} and PMC-VQA~\citep{zhang2023pmcvqa} fine-tune on figure-caption pairs from PubMed Central but discard the referring prose and cross-element context that ground scientific reasoning, resulting in poor performance in precise in-depth analysis. 

To close the gap, we present \sys, a system that automatically converts raw biomedical papers into an evidence-enriched training set and an enhanced specialized VLM. \sys improves post-training quality and efficiency through two mechanisms: (1)~it uses \emph{evidence-enriched prompting} to synthesize QA examples where each question is paired with the comprehensive supporting evidence from the source PDF, namely, the visual element, its caption, extracted structure, and the referring paragraphs. Also, it applies \emph{chart/table-aware OCR and structure extraction}, together with \emph{LLM-based cleansing}, to reduce layout- and OCR-induced errors before training. (2)~it adopts a \emph{progress-gated post-training strategy}: BioVLM automatically detects when additional synthesis and supervised fine-tuning stop improving benchmark performance, and then switches to reinforcement learning to strengthen evidence-grounded reasoning for complex biomedical question answering.

\sys shifts biomedical-related domain adaptation from an "expert-annotation bottleneck" to a reproducible pipeline that a small lab or team can run. Applying \sys to Qwen3-VL-8B produces BioVLM-8B, a strong and compact model that outperforms public, much larger and more expensive models such as GPT-5.2, while being post-trained end-to-end for under \$200. On challenging LAB-Bench (1,967 samples across eight biology categories), BioVLM-8B achieves \textbf{48.0\%} weighted accuracy, outperforming the strong GPT-5.2 by \textbf{3.8\,pp}, while remaining deployable on a single consumer GPU or Apple Silicon devices. Under an equal-token budget (a metric to reflect the post-training data efficiency), \sys's synthesized SFT data also surpasses human-curated PubMedQA and MedQA by \textbf{+17.1} and \textbf{+14.7}\,pp, respectively. 

Motivated by these results, we released \sys for early access to biomedical scientists and received consistently positive feedback on its practicality for day-to-day research tasks. Although this paper focuses on biomedicine, \sys is designed to generalize to scientific papers more broadly. Our early experiments are promising beyond biomedicine, and at the time of submission we are expanding support to climate change, geoscience, and civil engineering.

%% ============================================================
%%  SECTION 2: SYSTEM DESIGN
%% ============================================================
\section{\sys Design and Implementation}
\label{sec:system}

\subsection{System Overview}
\label{sec:glance}

\sys is released as a reproducible, low-cost workflow system tailored for biomedical scientists. Consider a biology researcher with a large collection of open-access papers on gene regulation. The researcher feeds these PDFs into \sys, together with a base VLM (Qwen3-VL-8B) and an evaluation benchmark (LAB-Bench). Without any manual annotation, \sys automatically:

\begin{enumerate}[leftmargin=*,label=\arabic*),topsep=1pt,itemsep=0pt,parsep=0pt,partopsep=0pt]
  \item Extracts structured text, figures, and tables using chart-aware OCR;
  \item Synthesizes millions of tokens of domain QA data via agentic retrieval-augmented generation;
  \item Post-trains the model with SFT followed by GRPO; and
  \item Evaluates and iteratively improves via weakness-driven data augmentation.
\end{enumerate}

The result is an enhanced model that can often surpass GPT-5.2 on biology benchmarks, while remaining lightweight enough to run locally on a single consumer GPU or an Apple device in the lab. Local deployment is also essential when some papers or internal reports are privacy-sensitive and cannot be uploaded to public model services (e.g., GPT-5.2, Gemini, or Claude).

\begin{figure}[t]
\centering
\includegraphics[width=\columnwidth]{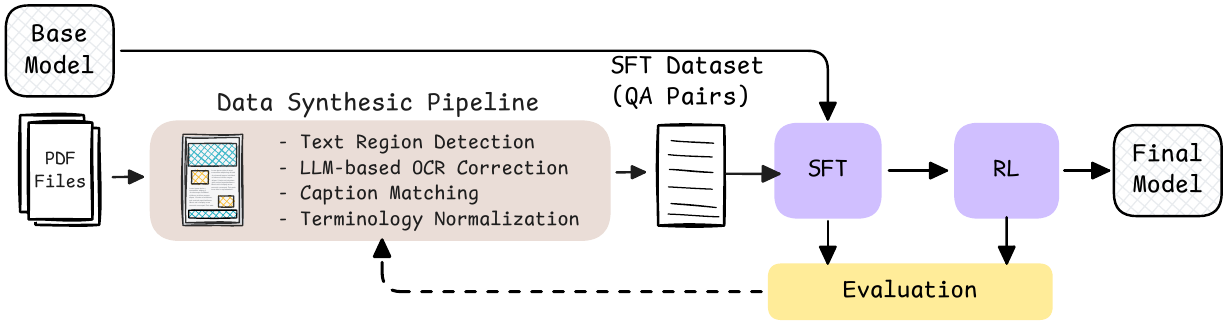}
\caption{The \sys pipeline overview. PDF files and a base model enter the Data Synthesis Pipeline, which trains the model via SFT then RL to produce the final domain-specialized model.}
\label{fig:pipeline}
\end{figure}

Figure~\ref{fig:pipeline} overviews the above pipeline.
\sys consists of four interconnected modules operating in a closed loop: document processing (\S\ref{sec:ocr}), data synthesis (\S\ref{sec:synthesis}), progress-gated training (\S\ref{sec:training}), and evaluation with feedback (\S\ref{sec:eval}).

\subsection{Chart-Aware Extraction and Cleansing}
\label{sec:ocr}

A common limitation for an automated data-synthesis pipeline is \emph{OCR hallucination}: if the extraction stage misreads a chart axis, drops a table row, or severs the link between a figure and its caption, every downstream QA pair inherits that error.
Layout-aware models such as LayoutLM~\citep{xu-etal-2021-layoutlmv2} and Nougat~\citep{blecher2023nougat} advance document extraction but target form-like documents or plain Markdown conversion without preserving cross-referential chains.
General-purpose tools such as Marker~\citep{paruchuri2025marker} and DeepSeek OCR~\citep{deepseek2025ocr} were designed for broad document conversion, not for the dense, multi-column layouts of scientific PDFs with embedded equations, multi-panel figures, and merged-cell tables.
In our early experiments, these tools produced frequent misrecognitions of gene names, chemical formulae, and chart annotations, where errors are propagated into training data and degraded downstream model accuracy.

\sys mitigates this with a five-stage extraction-and-cleansing pipeline:
(1)~\textbf{Layout detection} (Surya;~\citealp{paruchuri2025surya}) segments each page into text blocks, figures, tables, and captions, preserving spatial relationships that single-pass OCR discards.
(2)~Text regions are converted to structured Markdown preserving section hierarchy and paragraph boundaries.
(3)~\textbf{Cross-reference repair and content binding}: general-purpose OCR tools typically discard inline references (e.g., ``Table~1'', ``Figure~3'') during Markdown conversion, severing the link between body text and the visual elements it discusses. \sys recovers these cross-references, resolves figure/table labels, and binds each visual element with its caption \emph{and} the surrounding prose that interprets it, ensuring downstream synthesis can access the full evidential context.
(4)~Figures and tables receive \textbf{chart/table-aware extraction} (currently GLM-OCR;~\citealp{zaiorg2025glmocr}): charts are parsed into structured numerical descriptions; tables are converted to HTML with correct handling of merged cells and multi-row headers.
(5)~A \textbf{LLM-based cleansing stage} (currently Qwen3;~\citealp{qwen2025qwen3}) performs three passes---hallucination detection, domain-terminology repair (e.g., correcting ``BRAF V6OOE'' $\to$ ``BRAF V600E''), and cross-element consistency verification---before data enters the synthesis pipeline.
Because the preceding stages supply sufficient evidence and relevant context alongside each extracted element, we find that even compact models can perform cleansing reliably without introducing hallucinations.

The ablation in \S\ref{sec:ablation} quantifies the payoff: replacing our chart/table-aware extraction with generic alternatives degrades ChartQA accuracy by up to $-$7.8pp, and the gap widens monotonically with chart density (Table~\ref{tab:ocr}), confirming that scientific-domain-optimized extraction is necessary for high-fidelity evidence preservation.

\begin{figure}[t]
\centering
\includegraphics[width=\columnwidth]{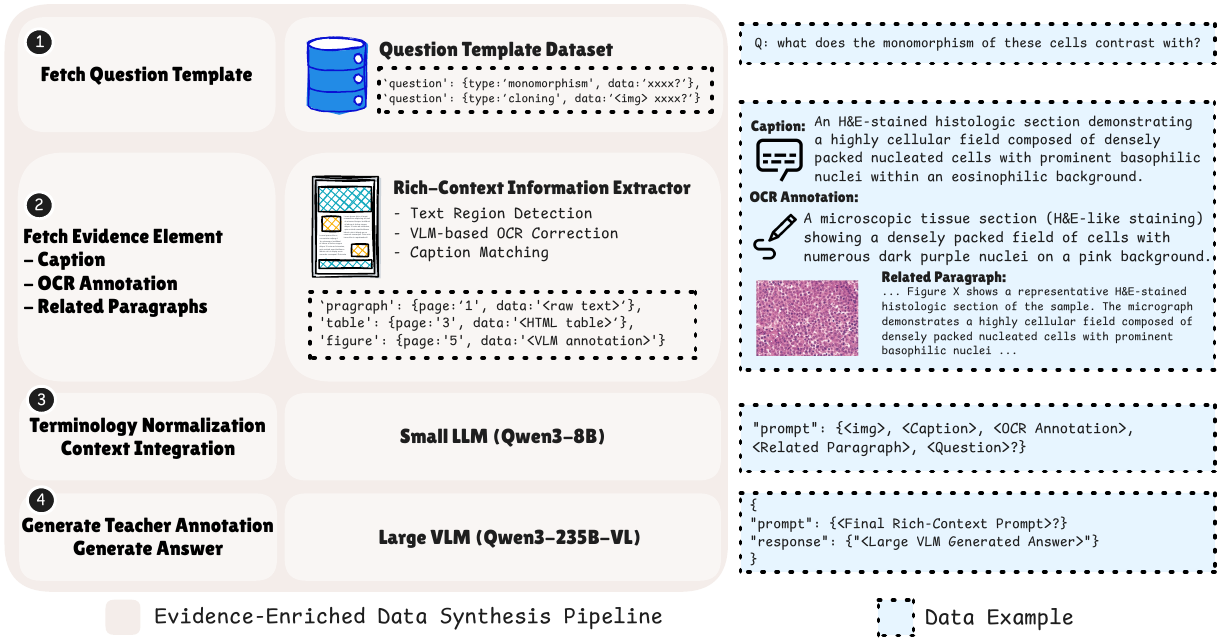}
\caption{The evidence-enriched data synthesis pipeline. Source PDFs undergo text region detection, LLM-based OCR correction, caption matching, and terminology normalization. The resulting evidence-enriched dataset preserves the full structural and semantic fidelity of the source material for downstream QA generation.}
\label{fig:architecture}
\end{figure}

\subsection{Evidence-Enriched QA Synthesis}
\label{sec:synthesis}

The synthesis pipeline transforms context-rich structured documents into the \textbf{evidence-enriched dataset}---high-quality QA training pairs that inherit the full structural and semantic fidelity of their source material.

\textbf{Task-Aware Question Generation.}
\sys draws question seeds from two complementary sources.
From \emph{raw papers}, the pipeline generates general domain questions (e.g., ``What is the function of protein X?'') and questions raised in different papers (e.g., ``Will X work on Mice?'').
From the \emph{evaluation benchmark}, the system identifies the \emph{reasoning skill categories} being tested (e.g., chart interpretation, protocol tracing, literature synthesis) rather than copying specific question templates.
These category-level cues are then extensively rephrased and diversified by a Qwen3-VL-235B model to produce new QA pairs whose surface forms are structurally distant from any benchmark item, with all answers grounded entirely in the source papers.
This process is analogous to curriculum-aware active learning: the benchmark informs \emph{which reasoning skills to prioritize}, not how to answer specific questions.
We emphasize that \textbf{no evaluation data---neither questions nor answers---is ever used in training}; all generated content derives exclusively from the source PDF corpus.
\textbf{On the boundary with benchmark contamination.}
A legitimate concern is whether this curriculum-aware approach constitutes indirect data leakage.
We argue it does not, for three reasons:
(1)~we extract only coarse-grained \emph{skill category labels} (e.g., ``chart interpretation'') from the benchmark taxonomy, never individual questions, answers, or answer distributions;
(2)~the same category labels could equally be derived from a textbook table of contents or a domain-expert syllabus---they carry no information specific to any benchmark item;
(3)~all generated answers are grounded exclusively in the source PDF corpus, which has zero overlap with LAB-Bench evaluation items.
Nonetheless, because the skill categories were selected \emph{with reference to} the target benchmark, performance gains on LAB-Bench should be interpreted as reflecting improved coverage of the reasoning skills that the benchmark is designed to test, rather than benchmark-independent generalization.
We encourage future work to evaluate on held-out benchmarks not referenced during data synthesis to further validate this distinction.

\textbf{Evidence-Retrieval Answer Generation.}
Each question is accompanied by evidence retrieved from the structured document store, utilizing content binding from the document processing stage (\S\ref{sec:ocr}) to enable precise extraction of passages, figures, and tables across document boundaries. All related elements in the paper are consolidated into the evidence set, including captions, OCR annotations (VLM annotations for charts and HTML format for tables), and associated paragraphs.

\textbf{Active Data Augmentation.}
The evaluation feedback loop (\S\ref{sec:eval}) identifies underperforming domain categories.
For each such category, \sys automatically searches for related open-access papers on PubMed, processes them through the document pipeline, and synthesizes additional QA pairs targeting the identified gaps.
Evaluation data is strictly excluded from all training; only the identified underperforming \emph{categories} guide the search for new source papers.

\subsection{Progress-Gated Training}
\label{sec:training}

\textbf{Intuition.}
Data synthesis is the most expensive stage of the pipeline (\S\ref{sec:cost}), yet more data does not always help: beyond a saturation point, additional SFT examples yield diminishing returns while RL over the same budget can unlock substantial gains.
\sys therefore separates training into a \emph{calibration} phase (SFT) that injects domain information and a \emph{refinement} phase (GRPO) that strengthens evidence-grounded reasoning, with an automatic \emph{progress gate} deciding when to switch.

\textbf{Progress Gate.}
After each {$\sim$}1M-token increment of synthesized data, \sys trains a checkpoint and evaluates on the target benchmark.
When accuracy improvement stagnates over consecutive rounds, \sys declares SFT saturation, freezes the dataset, and transitions to RL\@.

\textbf{Calibration Phase.}
SFT calibrates the base model to domain-specific knowledge via LoRA~\citep{zheng2024llamafactory}, alternating between text-only and vision QA batches so the model learns both domain terminology that appears in figures and textual descriptions that reference visual evidence.

\textbf{Refinement Phase.}
Once the progress gate is triggered, \sys converts the accumulated SFT data into an RL format and applies GRPO~\citep{hu2024openrlhf}, which trains the model to generate coherent reasoning chains without relying on a separate reward model. GRPO has been widely adopted for mathematical reasoning~\citep{shao2024deepseekmath}, where comparable reasoning demands exist; similarly, complex biomedical experimental results require not only domain knowledge in biology but also rigorous figure and chart interpretation. \sys integrates a progress-gated SFT-to-GRPO transition with evidence-enriched data for scientific document reasoning. During the SFT stage, it acquires common sense and fundamental biological concepts, leading to improvements in straightforward QA tasks such as SuppQA and DbQA, as shown in Table~\ref{tab:main_results}. During the RL stage, it achieves enhanced reasoning performance in domains that require more advanced analytical thinking.
As shown in \S\ref{sec:experiments}, SFT alone matches GPT-5.2 (43.7\% vs.\ 44.2\%), while GRPO provides the critical +4.3pp margin for overall superiority---confirming that calibration and refinement serve complementary roles.

\textbf{Token Budget and Quality Control.}
Rather than pre-specifying a fixed data budget, \sys incrementally grows the dataset in {$\sim$}1M-token increments, training and evaluating after each round.
When SFT performance plateaus, the system stops generation, converts the accumulated data into RL format, and switches to GRPO\@.
Throughout, LLM-based quality filtering and deduplication ensure data integrity.

\subsection{Evaluation and Feedback Loop}
\label{sec:eval}

The evaluation module assesses model performance across benchmark categories and performs \textbf{limitation detection} by identifying domains where performance falls below a configurable threshold.
Detected weaknesses update the QA template library with new question patterns targeting those domains, triggering a new round of data synthesis (\S\ref{sec:synthesis}).
This closed-loop design supports iterative refinement until performance converges.

%% ============================================================
%%  SECTION 4: EXPERIMENTS
%% ============================================================
\section{Evaluation}
\label{sec:experiments}

\subsection{Evaluation Setup}

We collect approximately 20{,}000 open-access biology papers from PubMed Central (PMC) as the source PDF corpus. We use Qwen3-VL-8B~\citep{qwen2025qwen3} as the base model. We evaluate on LAB-Bench~\citep{laurent2024labbench}, a biology research benchmark with 1{,}967 samples spanning eight diverse categories that test distinct reasoning skills---from literature comprehension (LitQA2) and figure/chart interpretation (FigQA, ChartQA) to sequence analysis (SeqQA), table reasoning (TableQA), and protocol understanding (ProtocolQA). Even state-of-the-art commercial models such as GPT-5.2 achieve only 44.2\%, confirming the benchmark's difficulty and discriminative power.
Results are reported as sample-weighted averages.

We include \textbf{GPT-5.1 mini} and \textbf{GPT-5.2} as commercial reference points.\footnote{GPT-5.2 results were obtained via the OpenAI API (\texttt{gpt-5.2-2025-12-11} checkpoint) in December 2025. LAB-Bench scores are not reported in OpenAI's blog post~\citep{openai2025gpt52}; all GPT-5.2 numbers in this paper are from our own evaluation runs.}
These represent particularly strong baselines for biomedical tasks: OpenAI positions the GPT-5 family as a flagship for healthcare applications, reporting state-of-the-art results on HealthBench~\citep{openai2025healthbench}---a physician-curated benchmark with 5{,}000 clinical scenarios---and achieving 92.4\% on GPQA Diamond graduate-level science questions including biology~\citep{openai2025gpt52}.
To validate that the evidence-enriched dataset is superior to existing alternatives, we train two baseline models using the same base model and training configuration:
(1)~\textbf{PubMedQA SFT}~\citep{jin2019pubmedqa} and
(2)~\textbf{MedQA SFT}~\citep{jin2021medqa}, both aligned to our token budget via truncation/duplication.

All training configurations use identical token budgets: 8{,}051{,}591 tokens for SFT and 1{,}584{,}412 tokens for GRPO, measured by the Qwen3-8B tokenizer. All experiments are conducted on a server with an AMD EPYC 7313P CPU and 4$\times$ NVIDIA RTX A6000 (48\,GB) GPUs.

\subsection{Main Results}

\begin{table}[t]
\centering
\caption{LAB-Bench results (accuracy \%). Qwen3-VL-8B (base) is the unmodified base model. BioVLM-8B models are trained from Qwen3-VL-8B. Best in \textbf{bold}.}
\label{tab:main_results}
\resizebox{\linewidth}{!}{%
\begin{tabular}{l|c|cc|cc|cc}
\toprule
 & Qwen3 & GPT & GPT & PubMed & MedQA & BioVLM-8B & \\
\textbf{Category} & VL-8B & 5.1m & 5.2 & SFT & SFT & (SFT only) & BioVLM-8B \\
\midrule
  Cloning    & 24.2 & 30.3 & 36.4 & 33.3 & 24.2 & 34.5 & \textbf{38.4} \\
  DbQA       & 31.2 & 40.6 & 41.7 & 14.0 & 20.0 & 44.7 & \textbf{48.9} \\
  FigQA      & 24.7 & 26.5 & \textbf{36.5} & 18.0 & 22.0 & 31.8 & 35.2 \\
  LitQA2     & 38.7 & 50.8 & 45.7 & 38.0 & 34.0 & 58.2 & \textbf{65.5} \\
  ProtocolQA & 38.3 & 63.0 & 65.7 & 46.0 & 46.0 & 68.1 & \textbf{72.3} \\
  SeqQA      & 43.4 & 36.2 & \textbf{47.0} & 34.0 & 34.0 & 39.5 & 42.8 \\
  SuppQA     & 24.8 & 36.6 & \textbf{48.8} & 30.0 & 30.0 & 40.9 & 44.2 \\
  TableQA    & 34.0 & 34.0 & 36.9 & 22.0 & 30.0 & 40.3 & \textbf{45.6} \\
\midrule
  \textbf{W.~Avg} & 35.4 & 39.1 & 44.2 & 26.6 & 29.0 & 43.7 & \textbf{48.0} \\
\bottomrule
\end{tabular}%
}
\end{table}

\textbf{BioVLM-8B surpasses GPT-5.2 by +3.8\,pp overall}, winning 5 of 8 categories, with the largest margins in LitQA2 (+19.8), TableQA (+8.7), and DbQA (+7.2)---categories that demand deep literature comprehension and structured data reasoning.
Compared to the unmodified Qwen3-VL-8B base model (35.4\%), BioVLM-8B achieves an absolute gain of +12.6\,pp.
BioVLM-8B (SFT only) alone matches GPT-5.2 (43.7\% vs.\ 44.2\%), while the GRPO stage provides the critical margin for overall superiority.
Equal-token human-curated datasets perform drastically worse (PubMedQA: 26.6\%, MedQA: 29.0\%), trailing BioVLM-8B (SFT only) by +17.1 and +14.7\,pp, confirming that the \emph{evidence-enriched dataset} is fundamentally superior to human-curated alternatives.
GPT-5.2 retains advantages in FigQA ($-$1.3), SeqQA ($-$4.2), and SuppQA ($-$4.6), which involve heavier visual understanding and sequence analysis.

\textbf{Cross-Model Generalization.}
To verify that the evidence-enriched dataset generalizes beyond one architecture, we apply the same SFT data to three additional base models (Table~\ref{tab:cross_model}).
All models show consistent improvement, with the largest gain on Qwen3-VL-8B (+12.6\,pp) which benefits from both text and vision components.

\begin{table}[t!]
\centering
\caption{Cross-model generalization: BioVLM-8B's evidence-enriched dataset improves all tested base models on LAB-Bench (weighted accuracy \%).}
\label{tab:cross_model}
\resizebox{0.9\linewidth}{!}{%
\begin{tabular}{l|ccc}
\toprule
\textbf{Base Model} & \textbf{Before SFT} & \textbf{After SFT} & \textbf{$\Delta$} \\
\midrule
Qwen2.5-7B       & 33.1 & 35.1 & +2.0 \\
LLaMA-3.2        & 31.3 & 34.4 & +3.2 \\
Gemma-2           & 31.8 & 33.5 & +1.7 \\
Qwen3-VL-8B      & 35.4 & 43.7 & +8.3 \\
\bottomrule
\end{tabular}%
}
\end{table}

\subsection{Ablation and Analysis}
\label{sec:ablation}

\textbf{OCR Method Comparison.}
To validate the necessity of chart-aware extraction, we compare four OCR configurations across three benchmarks of varying chart density (Table~\ref{tab:ocr}).
Replacing GLM-OCR with generic alternatives degrades ChartQA by up to $-$7.8pp (75.8\%$\to$68.0\%), with the gap widening as chart density increases: +7.8pp on ChartQA (pure charts), +1.65pp on SlideVQA (mixed), +0.51pp on ArXivQA (text-dominated).
This gradient confirms that specialized chart/table extraction is \emph{necessary} for high-quality evidence preservation.

\begin{table}[t!]
\centering
\caption{OCR pipeline comparison (accuracy \%). Our full document processing pipeline achieves the best performance, with larger gains on chart-intensive tasks.}
\label{tab:ocr}
\resizebox{0.9\linewidth}{!}{%
\begin{tabular}{l|ccc}
\toprule
\textbf{Method} & \textbf{ArXivQA} & \textbf{SlideVQA} & \textbf{ChartQA} \\
\midrule
Without OCR  & 72.61 & 64.83 & 68.0 \\
Marker       & 72.96 & 64.74 & 69.3 \\
DeepSeek OCR & 72.91 & 66.15 & 69.1 \\
\textbf{Ours} & \textbf{73.12} & \textbf{66.48} & \textbf{75.8} \\
\bottomrule
\end{tabular}%
}
\end{table}

\begin{figure}[t]
\centering
\includegraphics[width=\columnwidth]{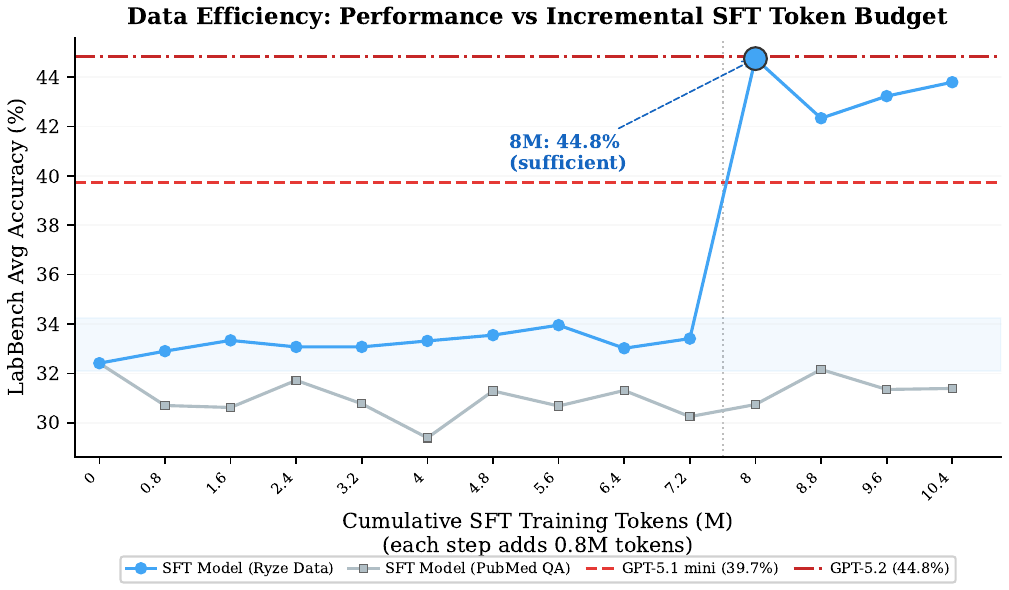}
\caption{Incremental SFT token budget vs.\ LAB-Bench accuracy. Dashed lines indicate GPT-5.1 mini and GPT-5.2 reference levels.}
\label{fig:token_ablation}
\end{figure}

\textbf{Token Budget Ablation.}
We incrementally grow the SFT dataset in {$\sim$}0.8M-token steps and evaluate after each round (Figure~\ref{fig:token_ablation}).
During the first 10 steps (0--7.2M tokens), performance plateaus at {$\sim$}33\% with $<$2pp variation---the model acquires frequency-level familiarity with domain terminology but has not yet deeply internalized the knowledge.
At 8M tokens, a sharp jump to 44.8\% occurs, already matching GPT-5.2.
Further tokens (8.8--10.4M) remain in the 42--44\% range, confirming that knowledge consolidation requires sufficient data volume, and additional data beyond the sufficiency point yields diminishing returns.

\begin{figure}[t]
\centering
\begin{minipage}[t]{0.54\columnwidth}
\centering
\includegraphics[width=\linewidth]{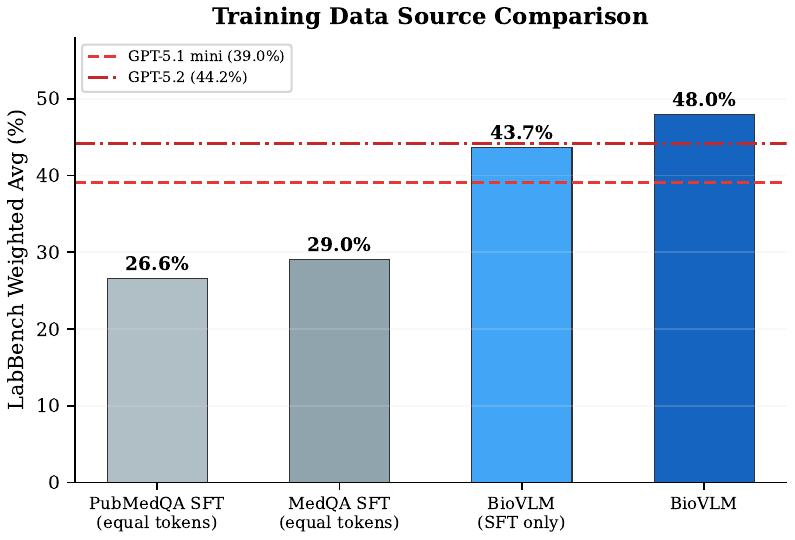}
\label{fig:datasource}
\end{minipage}
\hfill
\begin{minipage}[t]{0.43\columnwidth}
\centering
\includegraphics[width=\linewidth]{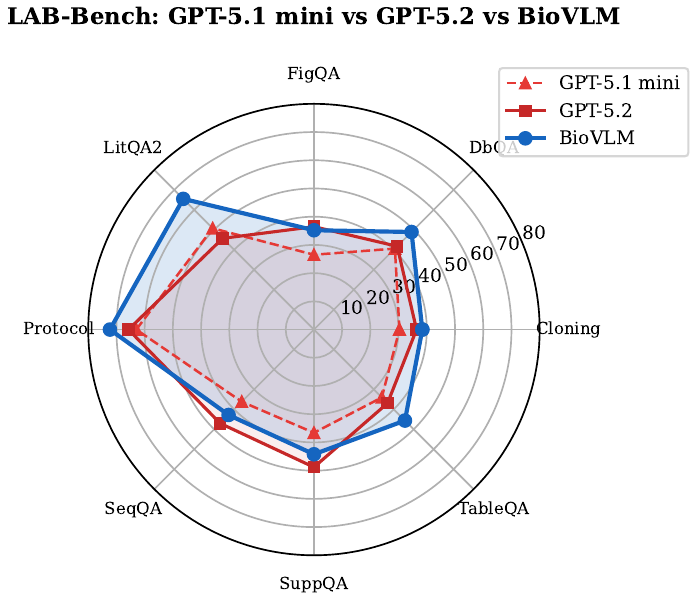}
\label{fig:radar}
\end{minipage}
\caption{Data source comparison and per-category analysis on LAB-Bench. \textbf{Left}: Weighted accuracy by data source under an equal 8.05M-token SFT budget; dashed lines indicate GPT-5.1 mini and GPT-5.2 reference levels. \textbf{Right}: Per-category radar profile of GPT-5.1 mini, GPT-5.2, and BioVLM-8B.}
\label{fig:datasource_radar}
\end{figure}

\textbf{Data Source Comparison.}
Under the same 8.05M-token SFT budget, PubMedQA achieves 26.6\% and MedQA reaches 29.0\%, while BioVLM-8B (SFT only) attains 43.7\%---a margin of +17.1 and +14.7pp respectively (Figure~\ref{fig:datasource_radar}).
With GRPO, BioVLM-8B further climbs to 48.0\%, surpassing GPT-5.2 by +3.8pp.
This controlled comparison isolates data quality as the decisive factor: the evidence-enriched dataset produced by \sys is fundamentally more effective than repurposing existing biomedical QA datasets.

\textbf{Progress-Gated Training Analysis.}
Table~\ref{tab:main_results} shows the progressive enhancement from commercial baselines through BioVLM-8B's two training stages across all eight LAB-Bench categories.
GRPO gains are largest on reasoning-intensive categories: LitQA2 (+7.3pp over SFT, +19.8pp over GPT-5.2), ProtocolQA (+4.2pp), and DbQA (+4.2pp).
Categories where GPT-5.2 retains an edge---FigQA, SeqQA, SuppQA---involve heavier visual understanding and sequence analysis, suggesting future improvements in multimodal RL training.
The radar chart in Figure~\ref{fig:datasource_radar} further highlights the per-category profile.
This pattern confirms the ``know the facts, then know the reasons'' paradigm: SFT injects domain information that GRPO consolidates into knowledge through reasoning.

\subsection{Cost Analysis}
\label{sec:cost}

Table~\ref{tab:cost} breaks down the estimated cost of reproducing the full pipeline on cloud GPUs.
The QA Synthesis stage includes the cost of running Qwen3-VL-235B for question diversification (\S\ref{sec:synthesis}); all model inference is performed locally on rented GPUs and is accounted for within the listed GPU hours.
The total end-to-end cost is \textbf{under \$200}---comparable to a single month of a commercial API subscription and at least an order of magnitude cheaper than expert human annotation of similar scale.

\begin{table}[t!]
\centering
\caption{Estimated cost on cloud GPUs (RunPod rates: A100 80\,GB \$1.19/hr; A6000 48\,GB \$0.76/hr).}
\label{tab:cost}
\resizebox{0.85\linewidth}{!}{%
\begin{tabular}{l|ccr}
\toprule
\textbf{Stage} & \textbf{GPU Config} & \textbf{Hours} & \textbf{Cost} \\
\midrule
OCR + Cleansing   & 1$\times$A6000  & ${\sim}$24h & ${\sim}$\$18 \\
QA Synthesis      & 8$\times$A100   & ${\sim}$15h & ${\sim}$\$143 \\
SFT Training      & 4$\times$A6000  & ${\sim}$8h  & ${\sim}$\$24 \\
GRPO Training     & 4$\times$A6000  & ${\sim}$4h  & ${\sim}$\$12 \\
\midrule
\textbf{Total}    &                 &             & $\mathbf{<\$200}$ \\
\bottomrule
\end{tabular}%
}
\end{table}

\begin{figure}[t]
\centering
\includegraphics[width=\columnwidth]{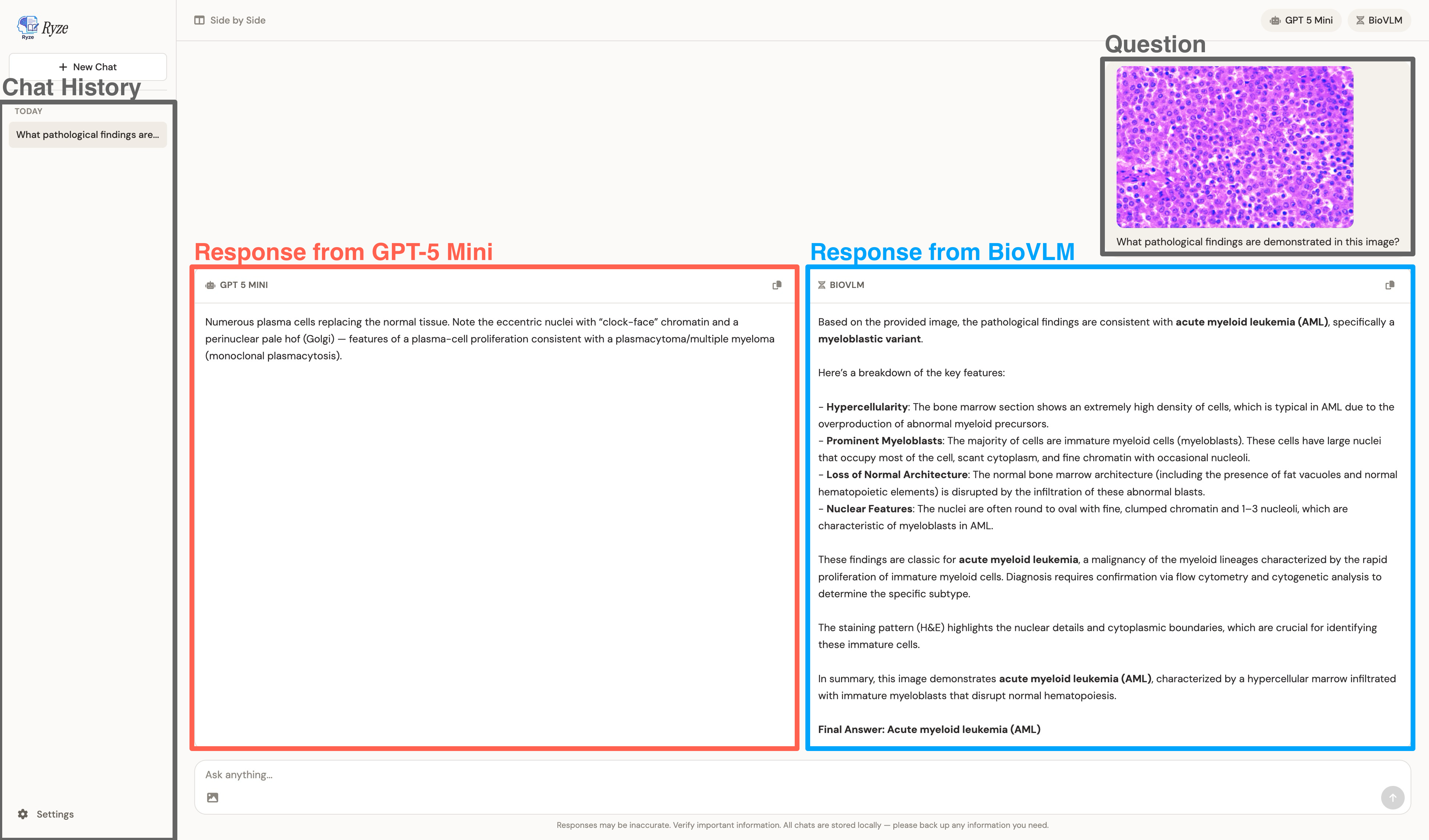}
\caption{The \sys demo interface, supporting side-by-side comparison between BioVLM-8B and commercial models such as GPT-5 Mini.}
\label{fig:demo}
\end{figure}

We release the complete pipeline code (Apache 2.0), evidence-enriched dataset, and BioVLM-8B model weights at \url{https://github.com/Chivier/Ryze}.
Figure~\ref{fig:demo} shows the \sys web demo, where users can query BioVLM-8B and compare its responses with commercial models.

%% ============================================================
%%  SECTION 5: CONCLUSION + LIMITATIONS
%% ============================================================
\section{Conclusion}

We present \sys, a fully automated system that converts raw biomedical papers into an evidence-enriched dataset and a domain-specialized VLM\@.
The core insight is that preserving the full contextual structure of scientific documents---figures, charts, tables, captions, and the prose that ties them together---produces training data that is fundamentally superior to both human-curated alternatives and context-agnostic automated approaches.
On LAB-Bench, BioVLM-8B achieves 48.0\% weighted accuracy, surpassing GPT-5.2 by +3.8\,pp and outperforming equal-token human-curated datasets by over 17\,pp, at an end-to-end cost under \$200.
We release the complete pipeline, dataset, and model to support domain-specialized training across scientific fields.

\section*{Limitations}

Our evaluation is limited to the biology domain; generalization to other scientific fields remains to be validated.
GPT-5.2 retains advantages in categories involving visual understanding (FigQA) and sequence analysis (SeqQA, SuppQA), indicating room for improvement in multimodal reasoning.
The scaling behavior of our progress-gated paradigm with larger models warrants further investigation.

We plan to address these limitations before presenting the demo at ACL, if accepted.

\bibliography{custom}

\end{document}